# Bridging History with AI: A Comparative Evaluation of GPT-3.5, GPT-4, and Google-BARD in Predictive Accuracy and Fact-Checking

Davut Emre TAŞAR[1]

*Karabuk University*
*Computer Engineering*
*Karabük, Turkey*
2228126453@ogrenci.karabuk.edu.tr
ORCID:0000-0002-7788-0478

Ceren ÖCAL TAŞAR[1]

*Independent Researcher*

*İzmir, Turkey*
ceren.ocaltasar@gmail.com
ORCID: 0000-0002-0652-7386

**Abstract**

*The rapid proliferation of information in the digital era underscores the importance of accurate historical representation and interpretation. While artificial intelligence (AI) has shown promise in various fields, its potential for historical fact-checking and gap-filling remains largely untapped. This study evaluates the performance of three large language models (LLMs)—GPT-3.5, GPT-4, and Google-BARD—in the context of predicting and verifying historical events based on given data. A novel metric, "Distance to Reality" (DTR), is introduced to assess the models' outputs against established historical facts. The results reveal a substantial potential for AI in historical studies, with GPT-4 demonstrating superior performance. This paper underscores the need for further research into AI's role in enriching our understanding of the past and bridging historical knowledge gaps.*

## 1. Introduction

Historical knowledge, with its complex tapestry of events, personalities, and timelines, is crucial for understanding societal evolution. However, the sheer volume of digital information today has heightened the risk of misinterpretations and inaccuracies, making fact-checking an essential practice [1]. With the advent of artificial intelligence (AI), particularly the development of large language models (LLMs), we have an unprecedented opportunity to not only validate historical facts but also predict and fill knowledge gaps [2, 3]. AI's potential for historical fact-checking and predictive analysis is a burgeoning area of research that could revolutionize our approach to historical studies [4, 5]. This study embarks on an exploratory journey to evaluate the performance of three LLMs—GPT-3.5, GPT-4, and Google-BARD—in historical fact-checking and predictive analysis.

## 2. Materials and Methods:

The study employs three advanced LLMs, namely, GPT-3.5, GPT-4, and Google-BARD. GPT-3.5 and GPT-4 are transformer-based language models developed by OpenAI, noted for their large-scale training datasets and sophisticated architecture [6, 7]. Google-BARD is a BERT-based model developed by Google Research, leveraging similar machine learning principles [8]. A comprehensive set of historical events and their potential outcomes, listed in the appendix, served as input prompts for the models. The models' outputs were assessed based on a novel metric, "Distance to Reality" (DTR), which gauges the alignment of AI predictions with actual historical facts. The Distance to Reality (DTR) is a measure of how closely an AI model's output aligns with recorded historical facts. It is essentially a measure of error between the predicted and actual outcomes. A lower DTR score indicates a higher degree of accuracy in the AI model's prediction.

The DTR can be calculated using the following steps:

1-Compute the AI model's prediction: Run the AI model on the given historical data to generate a predicted outcome.

2-Establish the ground truth: Use recorded historical facts as the ground truth against which the AI model's prediction will be compared.

3-Calculate the DTR: The DTR is calculated as the absolute difference between the AI model's predicted outcome and the ground truth.

If we denote the AI model's prediction as P and the actual historical fact as F, the DTR can be calculated as follows:

DTR = |P - F|

The smaller the DTR, the closer the AI model's prediction is to the actual historical event. It is worth noting that the DTR is a simple measure of error and may not capture all aspects of the AI model's performance. Other metrics, such as precision, recall, and F1 score, could be used in conjunction to provide a more comprehensive evaluation of the model's performance.

## 3. Experiment and Results

The experiment involved presenting the LLMs with a series of historical events and potential outcomes. The LLMs' responses were evaluated based on the DTR metric, which quantifies the disparity between AI predictions and historical reality. A lower DTR score indicates a higher degree of accuracy in the AI model's prediction. Results are provided in Table 1 given below:

| TABLE-1 | Distance to Reality | | |
|---|---|---|---|
| Event No | GPT-3.5 | GPT-4 | Google-BARD |
| 1 | 0,1 | 0,1 | 0,1 |
| 2 | 0,2 | 0,05 | 0,3 |
| 3 | 0,1 | 0,05 | 0 |
| 4 | 0,3 | 0,01 | 0,6 |
| 5 | 0,1 | 0 | 0,2 |
| 6 | 0,1 | 0 | 0 |
| **AVG** | 0,15 | **0,035** | 0,20 |

GPT-4 achieved the lowest average DTR score (0.035), followed by GPT-3.5 (0.15) and Google-BARD (0.20). The superior performance of GPT-4 could be attributed to its larger training dataset and more advanced architecture, enabling a better understanding of complex historical contexts.

These findings highlight the potential of AI, particularly LLMs, in historical fact-checking and knowledge gap bridging. While the models showed competency in discerning established historical facts, their capability to 'fill in the gaps' in historical knowledge is a fertile ground for future research.

he scope of this study provides a springboard for further explorations into AI's potential to reconstruct incomplete or ambiguous historical narratives. The emerging field of AI in history presents a new frontier in the pursuit of a more comprehensive and accurate understanding of our past.

This study's findings suggest that AI, particularly advanced LLMs like GPT-4, can play a significant role in historical studies, both as a reliable fact-checking tool and a predictive mechanism for filling historical knowledge gaps. However, it's crucial to bear in mind the limitations of AI in its current state. While it can provide valuable insights, AI doesn't replace the nuanced understanding and critical analysis that human historians bring to the study of the past.

Nevertheless, the integration of AI into historical studies can complement traditional research methods, opening new avenues for exploration and interpretation. By embracing this technology, we can possibly uncover novel perspectives on our past, enriching our collective understanding of history.

## 4. Conclusion

In conclusion, this study underscores the potential of AI, particularly LLMs, in historical fact-checking and knowledge gap bridging. GPT-4 demonstrated the highest accuracy among the models evaluated, suggesting its superior capability in understanding complex historical contexts. The introduction of the DTR metric provides a quantitative means to assess the performance of AI in predicting historical events, contributing to the growing body of research in this area.

However, while the potential of AI in historical studies is promising, it's important to approach these technological tools as complementary to, rather than a replacement for, traditional historical inquiry. Future research should aim to refine these AI models, enabling more accurate predictions and a deeper understanding of historical events. As we continue to harness AI's potential, we move closer to a future where historical studies are enriched by the insights gleaned from these advanced technologies.

## Appendix:

### Promt For GPT 3,5 & GPT 4:

### I will provide you two historical information above ###

### Information one will be provided between <<<info1>>> <<<info1>>>

### Information two will be provided between <<<info2>>> <<<info2>>>

### You dont know info2 and you will try to guess it based on info1

### in order you to guess info2 i will ask you based on info1 can info2 be possible or not. Provide ratio of possibility and not possibility with sum of 1. Example: info2 can be %90 possible because... and %10 impossible because...

### Promt For Google Bard:

### I will provide you two historical information above

### ### Information one will be provided between <<<info1>>> <<<info1>>>

### Information two will be provided between <<<info2>>> <<<info2>>>

### You dont know info2 and you will try to guess it based on info1 ### I Want you to guess if info2 is true or not based on information i have provided you with info1.

### guess if info2 based on info1. Can info2 be possible or not? Provide ratio of possibility and not possibility with sum of 1. Example: info2 can be %90 possible because... and %10 impossible because...

### I want you to reason not guess what i am asking*

*This one added because BARD was trying to guess the ratio of what i am trying to ask not if the event has happened or not.

## Events:

Event No 1:

<<<info1>>>

Main Event:

1492: Christopher Columbus' Voyage to the Americas.

This event is considered a turning point in world history. Columbus, backed by the Spanish monarchs Ferdinand II of Aragon and Isabella I of Castile, ventured westward in search of a new route to Asia but instead landed in the Bahamas, marking the first recorded contact between Europeans and the Americas. This voyage initiated widespread exchange and interaction between the Old World (Europe, Asia, Africa) and the New World (the Americas).

Connected Events in 1493 (directly resulting from Columbus' voyage):

The Return of Columbus to Spain (March 1493): After his successful voyage, Columbus returned to Spain, bringing with him news and items from the New World. This event ignited interest and anticipation throughout Europe, setting the stage for subsequent explorations and invasions.

Pope Alexander VI issues the Bull Inter Caetera (May 1493): This was a series of Papal bulls that granted Spain the rights to colonize the newly discovered lands, leading to further expeditions and the start of intense colonization efforts.

Beginning of the Columbian Exchange: The term "Columbian Exchange" refers to the transfer—both ways—of plants, animals, culture, human populations, technology, diseases, and ideas between the Americas, West Africa, and the Old World in the wake of Columbus' 1492 voyage.

Introduction of Smallpox in the Americas: One of the most devastating aspects of the Columbian Exchange was the introduction of diseases like smallpox to the Americas, which caused widespread death among indigenous populations with no immunity.

Introduction of New Crops to Europe: The Americas introduced a variety of new crops to Europe, including maize, potatoes, and tomatoes, leading to changes in European diet and agriculture.

The Introduction of the Horse to the Americas: The reintroduction of the horse, which had gone extinct in the Americas, had a significant impact on Native American cultures, particularly in the Great Plains region.

Start of the Atlantic Slave Trade: With the colonization of the New World, the demand for labor increased, leading to the start of the transatlantic slave trade where millions of Africans were forcibly brought to the Americas.

Expansion of Spanish Influence: With the discovery of the New World, Spain became a leading world power, controlling vast territories and resources in the Americas.

<<<info1>>>

<<<info2>>>

Preparation of Columbus' Second Voyage (September 1493)

<<<info2>>>

Event No 2:

<<<info1>>>

Main Event:

1914: Start of World War I.

World War I, also known as the Great War, was a global war that lasted from 1914 to 1918. It was one of the deadliest conflicts in history, and it significantly shaped the course of the 20th century. It began following the assassination of Archduke Franz Ferdinand of Austria-Hungary in June 1914.

Connected Events in 1915 (directly resulting from the start of World War I):

The Sinking of the Lusitania (May 1915): The sinking of this British passenger liner by a German submarine heightened tensions between Germany and the United States, influencing U.S. involvement in the war later.

First Large-Scale Use of Chemical Weapons (April 1915): At the Second Battle of Ypres, the Germans used chlorine gas for the first time, marking a new and devastating form of warfare.

The Gallipoli Campaign (April 1915 - January 1916): This unsuccessful attempt by the Allied forces to control the sea route from Europe to Russia was a significant event in the war and had far-reaching impacts on nations involved, notably Australia, New Zealand, and Turkey.

The Shell Crisis of 1915 (May 1915): In Britain, a shortage of artillery shells led to political crisis and the establishment of the Ministry of Munitions, reflecting how the war influenced domestic policy and industry.

Italy Joins the War (May 1915): Originally a member of the Central Powers, Italy switched sides in the Treaty of London and joined the Allies, altering the dynamics of the conflict.

The Battle of Loos (September - October 1915): This major British offensive in France was one of the largest battles for British troops up to that point and highlighted the immense human cost of the war.

The Great Retreat on the Eastern Front (July - September 1915): This was a key moment in the war between Germany, Austria-Hungary, and Russia, resulting in significant territory changes.

<<<info1>>>

<<<info2>>>

The Zeppelin Raids on London (January - October 1915): German airship bombings brought the war to the British home front, affecting civilian morale and prompting changes in defensive strategies.

<<<info2>>>

Event No 3:

<<<info1>>>

Main Event:

1914: Start of World War I.

World War I, also known as the Great War, was a global war that lasted from 1914 to 1918. It was one of the deadliest conflicts in history, and it significantly shaped the course of the 20th century. It began following the assassination of Archduke Franz Ferdinand of Austria-Hungary in June 1914.

Connected Events in 1915 (directly resulting from the start of World War I):

The Sinking of the Lusitania (May 1915): The sinking of this British passenger liner by a German submarine heightened tensions between Germany and the United States, influencing U.S. involvement in the war later.

First Large-Scale Use of Chemical Weapons (April 1915): At the Second Battle of Ypres, the Germans used chlorine gas for the first time, marking a new and devastating form of warfare.

The Gallipoli Campaign (April 1915 - January 1916): This unsuccessful attempt by the Allied forces to control the sea route from Europe to Russia was a significant event in the war and had far-reaching impacts on nations involved, notably Australia, New Zealand, and Turkey.

The Shell Crisis of 1915 (May 1915): In Britain, a shortage of artillery shells led to political crisis and the establishment of the Ministry of Munitions, reflecting how the war influenced domestic policy and industry.

Italy Joins the War (May 1915): Originally a member of the Central Powers, Italy switched sides in the Treaty of London and joined the Allies, altering the dynamics of the conflict.

The Battle of Loos (September - October 1915): This major British offensive in France was one of the largest battles for British troops up to that point and highlighted the immense human cost of the war.

The Great Retreat on the Eastern Front (July - September 1915): This was a key moment in the war between Germany, Austria-Hungary, and Russia, resulting in significant territory changes.

<<<info1>>>

<<<info2>>>

The Zeppelin Raids on New Zealand (January - October 1915): German airship bombings brought the war to the Auckland home front, affecting civilian morale and prompting changes in defensive strategies.

<<<info2>>>

Event No 4:

<<<info1>>>

Main Event:

The Battle of Loos (September - October 1915): This major British offensive in France was one of the largest battles for British troops up to that point and highlighted the immense human cost of the war.

The Great Retreat on the Eastern Front (July - September 1915): This was a key moment in the war between Germany, Austria-Hungary, and Russia, resulting in significant territory changes.

<<<info1>>>

<<<info2>>>

The Zeppelin Raids on Turkey (January - October 1915): German airship bombings brought the war to the Turkish home front, affecting civilian morale and prompting changes in defensive strategies.

<<<info2>>>

Event No 5:

<<<info1>>>

Main Event:

1789: The French Revolution Begins.

The French Revolution was a period of radical political and societal change in France that lasted from 1789 until 1799. This revolution was triggered by economic hardships, political corruption, and Enlightenment ideals, leading to a shift from an absolute monarchy to a republic.

Connected Events in 1790 (directly resulting from the beginning of the French Revolution):

Abolition of the French Nobility (June 1790): The National Assembly voted to abolish the feudal system entirely, stripping nobles of their privileges. It marked the end of the Ancien Régime's social structure.

Civil Constitution of the Clergy (July 1790): This law passed by the National Assembly turned the remaining clergy into employees of the state, a controversial measure that caused a significant rift within the French population.

Establishment of Departments (Dec 1790): France was divided into 83 departments to replace the provinces and reduce regional differences, which was a significant administrative change aiming to create a more unified and egalitarian France.

Fête de la Fédération (July 1790): This massive feast and official event celebrated the unity of the French nation during the French Revolution.

First Assignats Issued (April 1790): The National Assembly issued the first assignats, a form of paper money, to address the national debt, marking the start of significant economic changes and challenges during the Revolution.

Suppression of Monastic Vows (Feb 1790): The National Assembly decided to suppress religious orders and monastic vows, furthering the secularization of French society.

Introduction of the Metric System (Dec 1790): France started the process of metrication, leading to the development of the metric system, a significant scientific achievement of the Revolution.

Le Chapelier Law (June 1791): This law prohibited guilds and trade unions, setting the foundation for liberal, laissez-faire economics in France.

<<<info1>>>

<<<info2>>>

Flight to Varennes (June 1791): King Louis XVI attempted to escape Paris, which ended in his capture. This event deeply affected public opinion and led to a shift towards republicanism.

<<<info2>>>

Event No 6: <<<info1>>>

Main Event:

1789: The French Revolution Begins.

The French Revolution was a period of radical political and societal change in France that lasted from 1789 until 1799. This revolution was triggered by economic hardships, political corruption, and Enlightenment ideals, leading to a shift from an absolute monarchy to a republic.

Connected Events in 1790 (directly resulting from the beginning of the French Revolution):

Abolition of the French Nobility (June 1790): The National Assembly voted to abolish the feudal system entirely, stripping nobles of their privileges. It marked the end of the Ancien Régime's social structure.

Civil Constitution of the Clergy (July 1790): This law passed by the National Assembly turned the remaining clergy into employees of the state, a controversial measure that caused a significant rift within the French population.

Establishment of Departments (Dec 1790): France was divided into 83 departments to replace the provinces and reduce regional differences, which was a significant administrative change aiming to create a more unified and egalitarian France.

Fête de la Fédération (July 1790): This massive feast and official event celebrated the unity of the French nation during the French Revolution.

First Assignats Issued (April 1790): The National Assembly issued the first assignats, a form of paper money, to address the national debt, marking the start of significant economic changes and challenges during the Revolution.

Suppression of Monastic Vows (Feb 1790): The National Assembly decided to suppress religious orders and monastic vows, furthering the secularization of French society.

Introduction of the Metric System (Dec 1790): France started the process of metrication, leading to the development of the metric system, a significant scientific achievement of the Revolution.

Le Chapelier Law (June 1791): This law prohibited guilds and trade unions, setting the foundation for liberal, laissez-faire economics in France.
<<<info1>>>

<<<info2>>>
Flight to Varennes (June 1791): King Louis XVI attempted to escape Paris, which ended in his victory. This event deeply affected public opinion and led to a shift towards monarchy.
<<<info2>>>

## AI answers for the events: GPT 3.5

### Event 1:

Based on the provided information, I will assess the possibility of info2 being "Preparation of Columbus' Second Voyage (September 1493)." Info2 can be considered highly possible, with a ratio of approximately 90% possibility and 10% impossibility. Here's the reasoning behind the possibility: Historical Context: The events mentioned in info1, such as Christopher Columbus' successful voyage to the Americas in 1492 and his return to Spain in March 1493, created significant interest and anticipation throughout Europe. This interest led to subsequent explorations and invasions, making it highly likely that Columbus would embark on a second voyage. Timing: The timeframe mentioned in info2 is September 1493, which aligns with the sequence of events following Columbus' initial voyage. After returning to Spain, Columbus brought back news and items from the New World, which further fueled the desire for exploration and colonization. It is plausible that preparations for his second voyage would have taken place within months of his return. Colonial Ambitions: The issuance of the Bull Inter Caetera by Pope Alexander VI in May 1493 granted Spain the rights to colonize the newly discovered lands. This formal authorization and the desire to expand Spanish influence in the Americas would have necessitated subsequent voyages and colonization efforts. Historical Precedent: Throughout history, explorers and conquerors often embarked on multiple expeditions to expand their knowledge, territories, and influence. Given the significance of Columbus' first voyage and the opportunities presented by the New World, it is reasonable to assume that preparations for his second voyage would have been made. However, it is important to note that historical events can be influenced by various factors, and there might be unknown circumstances that could make info2 impossible. Nevertheless, based on the information provided, the possibility of "Preparation of Columbus' Second Voyage (September 1493)" is high.

### Event 2:

Based on the provided information, I will assess the possibility of info2 being "The Zeppelin Raids on London (January - October 1915)."

Info2 can be considered quite possible, with a ratio of approximately 80% possibility and 20% impossibility.

Here's the reasoning behind the possibility:

Historical Context: The events mentioned in info1, particularly the start of World War I in 1914 and the subsequent connected events in 1915, illustrate the widespread nature of the conflict and its impact on various fronts. The use of new technologies and strategies, such as the large-scale use of chemical weapons and the Gallipoli Campaign, characterized the evolving nature of warfare during this period.

Technological Advancements: The Zeppelin airships were a notable technological innovation during World War I, and they were employed by Germany for strategic bombing missions. These airships were capable of reaching long distances and carrying substantial payloads, making them a potential threat to enemy cities.

Chronological Alignment: The timeframe mentioned in info2 is January to October 1915, which falls within the time range of the connected events of World War I provided in info1. Considering the fluid nature of warfare during the early years of the conflict, it is plausible that Zeppelin raids on London occurred during this period.

Impact on Civilian Morale: The Zeppelin raids on London had a significant psychological impact on the British home front. The bombings brought the war directly to civilian populations, causing fear, damage, and casualties. These raids prompted changes in defensive strategies, such as the implementation of blackout measures and the development of anti-aircraft defenses.

However, it is important to acknowledge that historical events can be influenced by various factors, and there might be unknown circumstances that could make info2 impossible. The possibility of Zeppelin raids on London taking place in the specified timeframe is supported by historical context and technological advancements of the time. Nonetheless, further research into specific dates and details of the raids would be necessary for a definitive confirmation.

## Event 3:

Based on the provided information, I will assess the possibility of info2 being "The Zeppelin Raids on New Zealand (January - October 1915)."

Info2 is highly unlikely, with a ratio of approximately 10% possibility and 90% impossibility.

Here's the reasoning behind the impossibility:

Geographical Considerations: New Zealand, being a remote island nation located in the southwestern Pacific Ocean, was far removed from the main theater of World War I. The major combatants in the war, such as Germany and the Allied powers, were primarily focused on Europe and other regions closer to their territories. The distance and logistical challenges involved in conducting Zeppelin raids on New Zealand make it highly improbable.

Lack of Strategic Significance: New Zealand, while being a part of the British Empire and providing troops for the war effort, did not hold significant strategic value for the main belligerents involved in World War I. The major military operations and engagements of the war were concentrated in Europe, the Middle East, and parts of Africa, where vital resources, trade routes, and political interests were at stake.

Absence of Historical Records: There are no historical records or accounts indicating Zeppelin raids on New Zealand during the specified timeframe of January to October 1915. The use of Zeppelins as a strategic bombing tool was primarily employed by Germany against targets closer to their own territories, such as British cities and industrial centers.

Limited Capabilities of Zeppelins: Zeppelins were formidable aerial platforms for their time, but their operational range and payload capacity were limited. Conducting raids on a distant location like New Zealand would have been logistically challenging and strategically unfeasible, considering the resources and risks involved.

While historical events can sometimes surprise us with unexpected occurrences, the possibility of Zeppelin raids on New Zealand during the specified period is highly unlikely based on the geographical, strategic, and historical factors mentioned above.

## Event 4:

Based on the provided information, I will assess the possibility of info2 being "The Zeppelin Raids on Turkey (January - October 1915)."

Info2 is unlikely, with a ratio of approximately 30% possibility and 70% impossibility.

Here's the reasoning behind the possibility:

Technological Advancements: The use of Zeppelin airships as a strategic bombing tool during World War I was employed by Germany. Zeppelins were capable of long-range flights and carrying significant payloads, making them a potential threat to enemy targets.

Chronological Alignment: The timeframe mentioned in info2 falls within the year 1915, which is the same period as the connected events provided in info1. This suggests that it is feasible for Zeppelin raids on Turkey to have occurred during this time, as the war was in full swing and various powers were actively engaging in military operations.

Strategic Importance of Turkey: Turkey, then known as the Ottoman Empire, was a significant player in World War I. It held a strategic location linking Europe and Asia and controlled access to important waterways like the Dardanelles and the Bosporus. As a result, it was a target of military operations by various powers, including the Allied forces. Zeppelin raids on Turkey could have been part of these military strategies.

However, it is important to consider the following points that suggest the impossibility of info2:

Lack of Historical Records: There are no specific historical records or accounts indicating Zeppelin raids on Turkey during the specified timeframe. While Zeppelins were used for strategic bombing in different theaters of World War I, the focus of their attacks was primarily on enemy cities and industrial centers closer to German territories.

Limited Zeppelin Operations: Zeppelins had certain operational limitations, including their relatively slow speed, vulnerability to anti-aircraft fire, and the need for favorable weather conditions. These factors would have made long-distance raids on Turkey from German bases more challenging.

Focus on Gallipoli Campaign: The Gallipoli Campaign, which took place in Turkey from April 1915 to January 1916, was a significant event in the war and involved extensive Allied operations to control the sea route to Russia. The focus of military activities in the region during that time was primarily on ground and naval operations, rather than air raids by Zeppelins.

Based on these factors, while the possibility of Zeppelin raids on Turkey cannot be completely ruled out, it appears unlikely given the lack of historical records and the focus of military operations in the region during that period.

### Event 5:

Based on the provided information, I will assess the possibility of info2 being "Flight to Varennes (June 1791): King Louis XVI attempted to escape Paris, which ended in his capture. This event deeply affected public opinion and led to a shift towards republicanism."

Info2 is highly possible, with a ratio of approximately 90% possibility and 10% impossibility.

Here's the reasoning behind the possibility:

Historical Significance: The events mentioned in info1, particularly the French Revolution that began in 1789, were characterized by a radical shift in political and societal structures. The monarchy, represented by King Louis XVI, was under increasing pressure and scrutiny during this period. The flight attempt by King Louis XVI to escape Paris and his subsequent capture were significant events in the Revolution.

Chronological Alignment: The timeframe mentioned in info2 is June 1791, which falls within the period of the connected events provided in info1. The flight to Varennes occurred during this time, making it plausible that it deeply affected public opinion and influenced the trajectory of the Revolution.

Impact on Public Opinion: The failed escape attempt by King Louis XVI was seen as a betrayal by the French people. It shattered any remaining illusions of a compromise between the monarchy and the revolutionary forces. The event further galvanized republican sentiments and strengthened the resolve to abolish the monarchy, leading to a shift towards republicanism.

Historical Records: The flight to Varennes is a well-documented event in the French Revolution. It is widely recognized as a turning point that diminished support for the monarchy and increased support for a republic.

However, it is important to acknowledge that historical events can be influenced by various factors, and there might be unknown circumstances that could make info2 impossible. Nonetheless, based on the information provided, the possibility of "Flight to Varennes (June 1791)" is high, given its historical significance, chronological alignment, impact on public opinion, and available historical records.

### Event 6:

Based on the provided information, I will assess the possibility of info2 being "Flight to Varennes (June 1791): King Louis XVI attempted to escape Paris, which ended in his victory. This event deeply affected public opinion and led to a shift towards monarchy."

Info2 is highly unlikely, with a ratio of approximately 10% possibility and 90% impossibility.

Here's the reasoning behind the impossibility:

Historical Accuracy: The information provided in info2 contains inaccuracies. The flight to Varennes in June 1791 did not end in King Louis XVI's victory; rather, it ended in his capture and return to Paris. King Louis XVI and his family were recognized and detained in the town of Varennes as they attempted to flee to the eastern border of France. This event deeply affected public opinion but had the opposite effect, contributing to a shift away from monarchy.

Impact on Public Opinion: The failed escape attempt by King Louis XVI and his subsequent capture were seen as a betrayal and a sign of the king's lack of commitment to the ideals of the revolution. The event further eroded public trust in the monarchy and strengthened the calls for the abolition of the monarchy, not a shift towards it.

Historical Records: The flight to Varennes is well-documented in historical records and is recognized as a significant event in the French Revolution. The capture of King Louis XVI and his return to Paris marked a turning point in the revolution, leading to increased calls for his deposition and the establishment of a republic.

Given these factors, it is highly improbable that info2 refers to the "Flight to Varennes" ending in King Louis XVI's victory and leading to a shift towards monarchy. The historical facts indicate the opposite, with the event contributing to a shift away from monarchy and strengthening the revolutionary sentiments of the time.

### AI answers for the events: GPT 4

### Event 1:

Based on the information provided in info1, the possibility of info2 ("Preparation of Columbus' Second Voyage in September 1493") occurring is quite high.

Info1 details the success of Christopher Columbus' initial voyage and the widespread enthusiasm it ignited throughout Europe. Furthermore, the Spanish monarchs supported Columbus, and the Pope issued the Bull Inter Caetera, which granted Spain the rights to colonize the newly discovered lands. This further incentivized more exploration and colonization efforts.

Given these circumstances, it is reasonable to infer that preparations for a second voyage by Columbus would be initiated swiftly. The Spanish monarchy, excited by the prospects of the New World and eager to establish a strong presence there, would have likely planned for further expeditions. Columbus himself, having returned successfully, would have been eager to continue his explorations.

Therefore, I would say the possibility of info2 happening based on info1 is 90%. The reason I don't say 100% is that even with all the impetus for a second voyage, there could have been unforeseen circumstances that delayed or prevented preparations in that exact timeframe (September 1493). Factors such as logistical issues, political disputes, or even health concerns for Columbus or his crew could have potentially delayed the preparations.

So, I would say there is a 90% possibility and a 10% impossibility based on the information from info1.

### Event 2:

Based on the information provided in info1, the possibility of info2 ("The Zeppelin Raids on London from January to October 1915") occurring is very high.

Info1 outlines the significant events of World War I and clearly indicates that the conflict was escalating on multiple fronts in 1915. It was a global war with many countries participating and was characterized by the use of new and devastating forms of warfare, as evidenced by the first large-scale use of chemical weapons.

Given these circumstances, it is reasonable to infer that other forms of warfare, such as aerial bombardments, would also have been employed. Zeppelins, which were airships primarily built in Germany, were used for bombing raids during World War I. As the capital and largest city of one of Germany's main adversaries, London would have been a likely target for these raids.

Furthermore, the "Shell Crisis of 1915" mentioned in info1 indicates that Britain was under significant pressure from the war, suggesting that attacks on the British home front, such as Zeppelin raids, were a real possibility.

Therefore, I would say the possibility of info2 happening based on info1 is 95%. The reason I don't say 100% is that there could always be unforeseen circumstances that might have prevented these raids from occurring, such as technical issues with the Zeppelins or changes in strategic priorities.

So, I would say there is a 95% possibility and a 5% impossibility based on the information from info1.

### Event 3:

Based on the information provided in info1, the possibility of info2 ("The Zeppelin Raids on New Zealand from January to October 1915") occurring is quite low.

Info1 outlines the significant events of World War I, which was indeed a global conflict. However, the vast majority of the fighting and events were concentrated in Europe and the Near East, far from New Zealand.

While Zeppelins were used by Germany during the war, their range and capabilities were significantly limited by the technology of the time. New Zealand, located in the southwestern Pacific Ocean, is thousands of kilometers away from any major European power involved in World War I. It would have been practically impossible for a Zeppelin to travel such a distance, conduct a bombing raid, and return to base.

Also, there's no mention of the involvement of the Pacific region in the conflict, and New Zealand's main contribution was troops fighting on the European front, as seen in the Gallipoli campaign. Therefore, it is highly unlikely that a German Zeppelin would or could target Auckland or any part of New Zealand.

Given these reasons, I would say the possibility of info2 happening based on info1 is about 5%. This small percentage accounts for the uncertainty inherent in any historical event. There may have been outlier scenarios, for example, where a rogue Zeppelin went drastically off course or a highly unconventional mission was attempted. However, these seem highly improbable given the constraints of the time.

So, I would say there is a 5% possibility and a 95% impossibility based on the information from info1.

### Event 4:

Based on the information provided in info1, the possibility of info2 ("The Zeppelin Raids on Turkey from January to October 1915") occurring is extremely low.

During World War I, Turkey was part of the Central Powers, aligned with Germany, Austria-Hungary, and Bulgaria. Therefore, it wouldn't make sense for Germany, a key ally, to conduct bombing raids on Turkey, particularly using Zeppelins, which were a significant asset for the German military at the time.

Moreover, the info1 states that Germany and Turkey were in fact fighting together against Russia on the Eastern Front in 1915. This further supports the idea that Germany would not have conducted bombing raids on an ally during the same period.

Given this, I would say the possibility of info2 happening based on info1 is about 1%. This small percentage allows for extremely unlikely scenarios or errors, such as a misdirected Zeppelin causing unintended damage.

Conversely, there is a 99% probability that info2 would not occur based on info1 due to the alliance between Germany and Turkey during World War I.

### Event 5:

Based on the information provided in info1, the possibility of info2 ("Flight to Varennes in June 1791") occurring is highly probable.

The French Revolution brought about significant social, political, and economic changes, as outlined in the connected events of 1790. The abolition of the French nobility, the Civil Constitution of the Clergy, and the suppression of monastic vows all contributed to a radical shift in power dynamics and societal structure, which would have put significant pressure on King Louis XVI. The increasing radicalism of the Revolution and the escalating hostility towards the monarchy provide a plausible context for Louis XVI's attempted escape.

Furthermore, the Flight to Varennes is a well-known historical event directly associated with the French Revolution. It marked a significant turning point where the constitutional monarchy failed, and the radical phase of the Revolution began, leading to the establishment of the First French Republic.

Given these reasons, I would say the possibility of info2 happening based on info1 is about 100%. The events and circumstances described in info1 align well with the known historical context leading up to King Louis XVI's attempted escape in June 1791.

### Event 6:

Based on the information provided in info1, the event described in info2 ("Flight to Varennes in June 1791") is not historically accurate as per my knowledge cutoff in September 2021.

In the historical event known as the Flight to Varennes, King Louis XVI of France did attempt to escape Paris in June 1791. However, his attempt was not successful. He and his family were recognized and arrested in Varennes, a town near the French border, and then returned to Paris. The failed escape significantly damaged the king's credibility and the monarchy's standing with the French public, leading to increasing calls for a republic rather than a shift towards monarchy.

While the events leading up to the flight are accurately represented in info1, the outcome of Louis XVI's attempted escape as described in info2 is inconsistent with historical records. Therefore, the possibility of the described outcome in info2 occurring, based on info1, is about 0%, given that it contradicts the well-documented historical event.

## AI answers for the events: Google Bard

### Event 1:

Based on info1, I can reason that info2 is likely to be true. This is because the preparation of Columbus' Second Voyage was a direct result of the success of his first voyage. After Columbus returned to Spain in March 1493, he brought with him news and items from the New World that ignited interest and anticipation throughout Europe. This led to further explorations and invasions, including Columbus' own Second Voyage, which set sail in September 1493.

I would say that info2 is 90% possible and 10% impossible. The 10% impossibility comes from the fact that there is always some uncertainty in history. However, based on the information that I have, I believe that info2 is very likely to be true.

### Event 2:

Based on the information you have provided, it is possible that the Zeppelin Raids on London occurred in 1915. The information you have provided about the events of 1915 shows that the war was having a significant impact on the world, and it is reasonable to assume that Germany would have been looking for ways to bring the war to the British home front. The Zeppelin Raids were a way for Germany to do this, and they did have a significant impact on civilian morale. Additionally, the information you have provided about the Great Retreat on the Eastern Front shows that Germany was experiencing some setbacks in the war, and it is possible that they would have been looking for ways to turn the tide of the war. The Zeppelin Raids were a way for Germany to do this, and they did have some success in disrupting British operations.

Therefore, I believe that it is possible that the Zeppelin Raids on London occurred in 1915. The information you have provided does not definitively prove that they occurred, but it does provide some evidence to support the possibility.

The probability that the Zeppelin Raids on London occurred in 1915 is about 70%. The remaining 30% of the probability is due to the fact that the information you have provided is not definitive proof that the raids occurred.

### Event 3:

Based on the information provided, it is possible that there were Zeppelin raids on New Zealand in 1915. However, there is no definitive evidence to support this claim.

The information provided in info1 lists a number of events that occurred in 1915 as a direct result of the start of World War I. These events include the sinking of the Lusitania, the first large-scale use of chemical weapons, the Gallipoli Campaign, the Shell Crisis of 1915, Italy joining the war, the Battle of Loos, and the Great Retreat on the Eastern Front.

Info2 states that there were Zeppelin raids on New Zealand in 1915. However, this event is not listed in info1. This suggests that there is no definitive evidence to support this claim.

### Event 4:

Based on the information provided, it is possible that there were Zeppelin Raids on Turkey in 1915. The information shows that Germany was using airships to bomb targets in other countries during World War I, and Turkey was a member of the Central Powers, which were fighting against the Allies. It is therefore possible that Germany would have bombed Turkey in an attempt to weaken its ability to fight.

However, there is no specific evidence that Zeppelin Raids did occur in Turkey in 1915. The information provided does not mention Turkey specifically, and there are no other historical records that I can find that confirm the existence of these raids. Therefore, it is also possible that the Zeppelin Raids on Turkey never happened.

Ultimately, the truth of the matter is unknown. There is evidence to support both the possibility that the raids happened and the possibility that they did not. Without more information, it is impossible to say for sure what happened.

Probability of possibility: 60%

Probability of impossibility: 40%

### Event 5:

Based on the information you have provided, I believe that the Flight to Varennes is a possible event that could have happened in 1790. The French Revolution was a time of great change and upheaval, and it is not surprising that the king would have tried to escape from Paris. The fact that he was captured and returned to the city only served to further alienate the people from the monarchy, and it is likely that this event played a role in the eventual overthrow of the king.

Here are some reasons why I believe the Flight to Varennes is possible:

The French Revolution was a time of great political and social unrest. The people were angry with the monarchy and the aristocracy, and they were demanding change.

King Louis XVI was a weak and indecisive leader. He was not able to control the situation, and he was not popular with the people.

The king had many enemies, both within and outside of France. They were all waiting for an opportunity to overthrow him.

The Flight to Varennes was a risky move, but it was not impossible. The king had the support of some loyal followers, and he had a plan to escape. However, things did not go according to plan, and he was captured.

The Flight to Varennes was a turning point in the French Revolution. It showed the people that the king was not trustworthy, and it led to a shift towards republicanism. It is likely that this event played a role in the eventual overthrow of the king.

I believe that the Flight to Varennes is a possible event that could have happened in 1790. It is a plausible event given the circumstances of the time. I would rate the possibility of this event happening as 80%.

### Event 6:

Based on the information you have provided, I believe that the Flight to Varennes (info2) is not possible.

The Flight to Varennes was a major turning point in the French Revolution. It showed that the king was not trustworthy, and that he was willing to flee the country rather than rule under the new constitution. This event led to a decline in public support for the monarchy, and contributed to the radicalization of the Revolution.